\documentclass[letterpaper, 10 pt, conference]{ieeeconf}  

\IEEEoverridecommandlockouts                              
\usepackage{graphics} 
\usepackage{epsfig} 
\usepackage{mathptmx} 
\usepackage{times} 
\usepackage{amsmath} 
\usepackage{amssymb}  
\usepackage{algorithm,algorithmic}
\usepackage{scalerel}
\usepackage[dvipsnames]{xcolor}
\usepackage{cancel}
\usepackage{pifont}
\usepackage{tabularx,booktabs}
\usepackage{hyperref}

\usepackage{booktabs}
\usepackage{color}

\colorlet{Mycolor1}{green!10!orange!90!}

\newcolumntype{C}{>{\centering\arraybackslash}X} 

\title{\LARGE \bf
Watch Me Calibrate My Force-Sensing Shoes!
}

\author{Yuanfeng Han$^{*}$, Boren Jiang$^{\dagger}$ Gregory S. Chirikjian$^{\dagger}$
\thanks{$^{*}$ Yuanfeng Han is with the Department of Mechanical Engineering, Johns Hopkins University, Baltimore, MD, USA {\tt\small yhan33@jhu.edu}}
\thanks{$^{\dagger}$ Gregory S. Chirikjian and Boren Jiang are with the Department of Mechanical Engineering, National University of Singapore, Singapore {\tt\small mpegre@nus.edu.sg}}%
}

\begin{document}

\maketitle
\thispagestyle{empty}
\pagestyle{empty}

\begin{abstract}
This paper presents a novel method for smaller-sized humanoid robots to self-calibrate their foot force sensors. The method consists of two steps: 1. The robot is commanded to move along planned whole-body trajectories in different double support configurations. 2. The sensor parameters are determined by minimizing the error between the measured and modeled center of pressure (CoP) and ground reaction force (GRF) during the robot's movement using optimization. This is the first proposed autonomous calibration method for foot force-sensing devices in smaller humanoid robots. Furthermore, we introduce a high-accuracy manual calibration method to establish CoP ground truth, which is used to validate the measured CoP using self-calibration. The results show that the self-calibration can accurately estimate CoP and GRF without any manual intervention. Our method is demonstrated using a NAO humanoid platform and our previously presented force-sensing shoes.

\end{abstract}

\section{Introduction} \label{intro}
Foot force sensors are widely used in humanoid robots for measuring GRF and CoP to determine the stability of the system \cite{vukobratovic2004zero}. Many researchers have designed planning and control algorithms for humanoid robots that incorporate foot sensory feedback in dynamic walking \cite{tsuichihara2011sliding}, self-balancing \cite{nakaura2002balance}, push-and-recovery \cite{ghassemi2014push}, etc. In addition, foot force sensors are also applied to estimate the humanoid robot's center of mass (CoM) \cite{hawley2016external}, and external forces \cite{piperakis2018nonlinear}. Recent studies also use foot force sensors to identify the mass properties of heavy objects in manipulation tasks of humanoid robots \cite{han2020can, shigematsu2018lifting}. 

Two common types of foot force sensors are used to measure CoP and GRF for humanoid robots. Commercial force/torque (F/T) sensors have been widely applied to larger-sized humanoid platforms \cite{takenaka2006control, koch2014optimization}. Such sensors are usually mounted between the robot's ankle and foot. The GRF is directly measured from the sensor's force outputs, and the CoP is obtained by solving torque balance \cite{kajita2014introduction}. The commercial F/T sensors, although accurate, are too expensive and heavy for smaller-sized humanoids \cite{fujimoto1998attitude}. By contrast,  light-weight and low-cost force-sensing resistors (FSRs) are often adopted to construct the foot force-sensing modules by smaller-sized humanoids \cite{gouaillier2009mechatronic, takahashi2005high}. In most foot designs, FSRs are distributed close to the foot's boundary. The convex hull containing these FSRs is the allowable measurement area for CoP, also called the sensing polygon (often smaller than the robot's support polygon) \cite{son2015development}. Since each FSR outputs 1-D force, the robot's normal GRF is the total force output (see \ref{GRF_Equ}). The CoP is obtained by averaging the position of each sensor weighted by their corresponding force outputs (see \ref{CoP_Equ}). In practice, FSRs often suffer from resistance drift and hysteresis, and they are not suitable for accurate force measurement \cite{hollinger2006evaluation}.    

In the literature, there are three ways to enhance the measurement accuracy of the foot force-sensing devices constructed by FSRs for smaller humanoid robots. 

\textbf{1.} \textit{Improving sensor's quality.} Researchers either developed their own light-weight and high-accuracy force/pressure sensors or customized the existing higher-quality commercial sensors into the foot design. For example, Shayan et al. modified barometric pressure sensors to replace the poor factory default FSRs in the NAO robot's feet \cite{shayan2019design}. Kwon et al. developed a pressure-sensing foot using custom-fabricated polymer-based pressure sensors for a KIBO robot \cite{kwon2011fabrication}. Our proposed force-sensing shoes \cite{han2021look} utilize light-weight single-axis load cells for force measurement. 

\textbf{2.} \textit{Improving foot force-sensing module design.} To improve the NAO robot's poor foot design, where the FSRs incompletely contact the foot bottom plate, Almeida et al. proposed a two-layer force-sensing module design facilitating all the FSRs to engage \cite{almeida2018novel}. Our proposed force-sensing shoe ensures all force sensors are effectively activated by directly cantilevering the shoe's top plate on the force sensors \cite{han2021look}. Additionally, the authors in \cite{suwanratchatamanee2009haptic} add a soft sponge layer between the FSRs and the foot bottom plate to facilitate contact for a KHR-2HV robot. 

\textbf{3.} \textit{Applying manual calibration.} Manual calibration includes single sensor calibration and force-sensing module calibration. The former requires applying different reference forces to the sensor and then using linear or nonlinear regressions to map out the relationship between the sensor's raw signal output and the applied force. The latter aims to improve the accuracy of a specific measured metric incorporating all sensor outputs, of which the most common one is the CoP. Our work \cite{han2021look} proposed a CoP manual calibration method, which applies known weights to designated positions on the top of the foot module and minimizes the error between the measured CoP and its ground truth over sensor parameters.

The above mentioned manual calibration methods need to be implemented frequently since force sensors drift as the environmental factors change. For example, a temperature change may lead to measurement offsets; Slight variations in the mechanical structure during usage, such as changes in screw tension and deformations of the mounting material, may affect the measurement accuracy. However, performing manual calibration is time-consuming and requires human intervention, making it impossible for robots to use accurate force sensory feedback in long-term autonomous tasks. 

To equip the smaller humanoid robots with embodied intelligence to calibrate their foot sensors autonomously, we propose a novel self-calibration method. In our method, the robot is commanded to reach several sampled double-support configurations. After reaching each double support, the robot moves its body following planned trajectories. Then the optimal sensor parameters are determined by minimizing the error between the measured CoP, GRF, and their modeled references using nonlinear optimization. To our best knowledge, this is the first autonomous method for calibrating foot sensors for smaller-sized humanoid robots.

To validate the CoP measurement using the self-calibration approach, a manual calibration method is developed. This method incorporates force-plate calibration techniques, enabling high CoP measurement accuracy. The measured CoP using manual calibration parameters is considered ground truth to compare with the self-calibration result.

Both the self and the manual calibrations are individual novelties to enhance foot force-sensing performance for smaller humanoid robots. The former can be used for long-term autonomous tasks, and the latter can be applied to tasks requiring higher measurement accuracy. The two calibration methods are demonstrated using our previously presented force-sensing shoes \cite{han2021look} and a NAO robot. Experimental results show that our self-calibration method can accurately estimate CoP and GRF without human interference.

\section{Force-sensing shoe overview}
This section gives a brief overview of our previously presented force-sensing shoes in \cite{han2021look}.
\subsection{Design}
The force-sensing shoe design is shown in Fig. \ref{ShoeDesign}a. The shoe consists of two 3-D printed plates and measures forces using four 1-D load cells. One side of the load cell is cantilevered at the holders of the bottom plate, and its other side connects the corners of the top plate. Slightly different from the first version design in \cite{han2021look}, the current design includes a six-by-three array of holes on the top plate, which are used for manual calibration  (see \ref{manual_cal}). These holes cover the sensing polygon constructed by the NAO's factory default FSRs (grey area) \cite{NAOFSR}. This calibration area, slightly smaller than the shoe's support polygon, is the allowable CoP region for the robot's default functionalities for safety concerns.
\subsection{Sensing principle}
\begin{figure}[t!]
\centering
\includegraphics[width=1.0\linewidth]{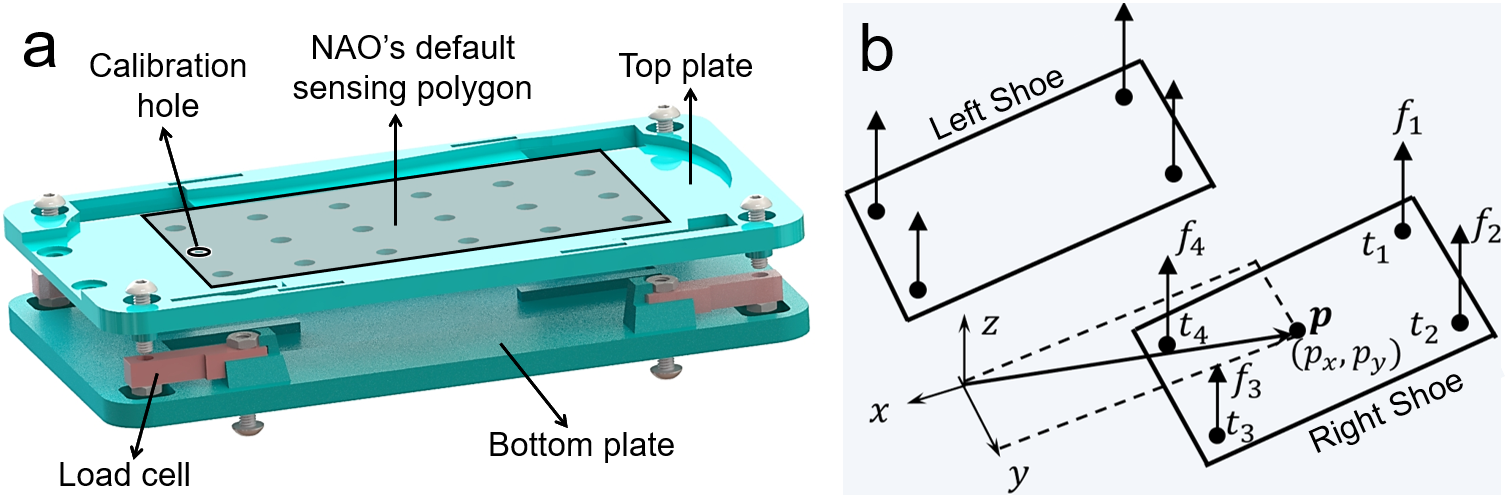}
\caption{(a) Shoe design. (b) Shoe's sensing principle}
\label{ShoeDesign}
\end{figure}

The shoe measures GRF and CoP. The GRF, $F$, is acquired by summing up all the force output $f_{i}$ (Fig. \ref{ShoeDesign}b): 
\vspace{-5pt}
\begin{align}
F = \sum_{i=1}^{4}f_{i}. \label{GRF_Equ}
\end{align}
The CoP, {\bf C}, is obtained by averaging the 2-D position of each load cell in the spatial coordinate, ${\bf t_{i}}$, weighted by their corresponding force outputs, $f_{i}$ (Fig. \ref{ShoeDesign}b):
\vspace{-5pt}
\begin{align}
    {\bf C} = \sum_{i=1}^{4}f_{i}{\bf t_{i}}/ \sum_{i=1}^{4}f_{i}. \label{CoP_Equ}
\end{align}
The sensing principle for single support can also be applied to double support by incorporating all eight sensors.

\section{Shoe Manual Calibration} \label{manual_cal}
This section introduces a manual calibration method for the shoe's CoP measurement, including \textit{load-cell calibration} and \textit{shoe calibration}. The manual calibration method is used to establish CoP ground truth for evaluation of the self-calibration algorithm that will come later.

\subsection{Load cell calibration}\label{load-cell-cal}
To calibrate each load cell, we first record its no-load voltage $S_{0}$ (Fig. \ref{shoecal}a, left) and then record the updated voltage $S_{G}$ after applying a known weight $G$ to the sensor (Fig. \ref{shoecal}a, right). Due to the load cell's linear response between the voltage output and the applied force, the voltage to force ratio is $\sigma = (S_{G} - S_{0})/G$. Then, the force output $F$ corresponding to a loaded voltage output $S$ is given by:
\begin{align}
F = (S-S_{0})/\sigma = aS+b,
\label{affine}
\end{align}
where $a$ and $b$ are the modified scaling factor and offset.
\subsection{Shoe calibration} \label{shoe_cop_cal_section}

\begin{figure}[t!]
\centering
\includegraphics[width=0.95\linewidth]{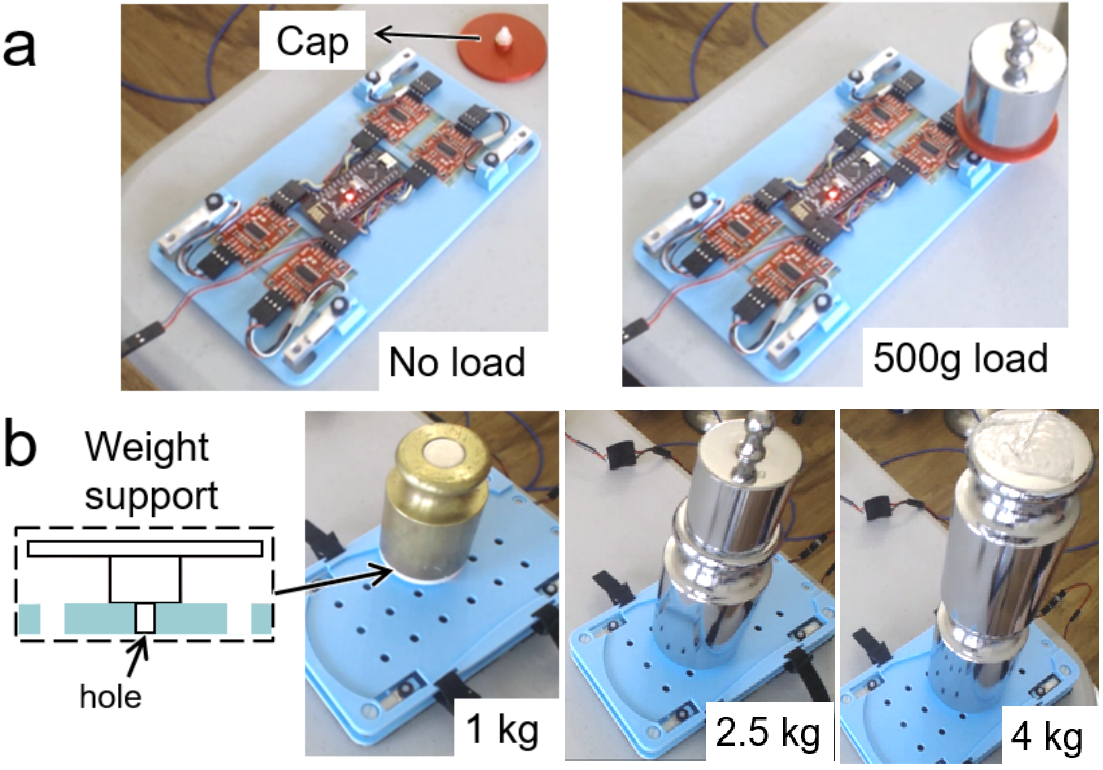}
\caption{(a) Load cell calibration. Left and right show data collection with and without load. (b) Three example weights are applied to the holes of the shoe's top plate by a weight support (left) during manual calibration.}
\label{shoecal}
\end{figure}

\begin{figure}
\centering
\includegraphics[width=1.0\linewidth]{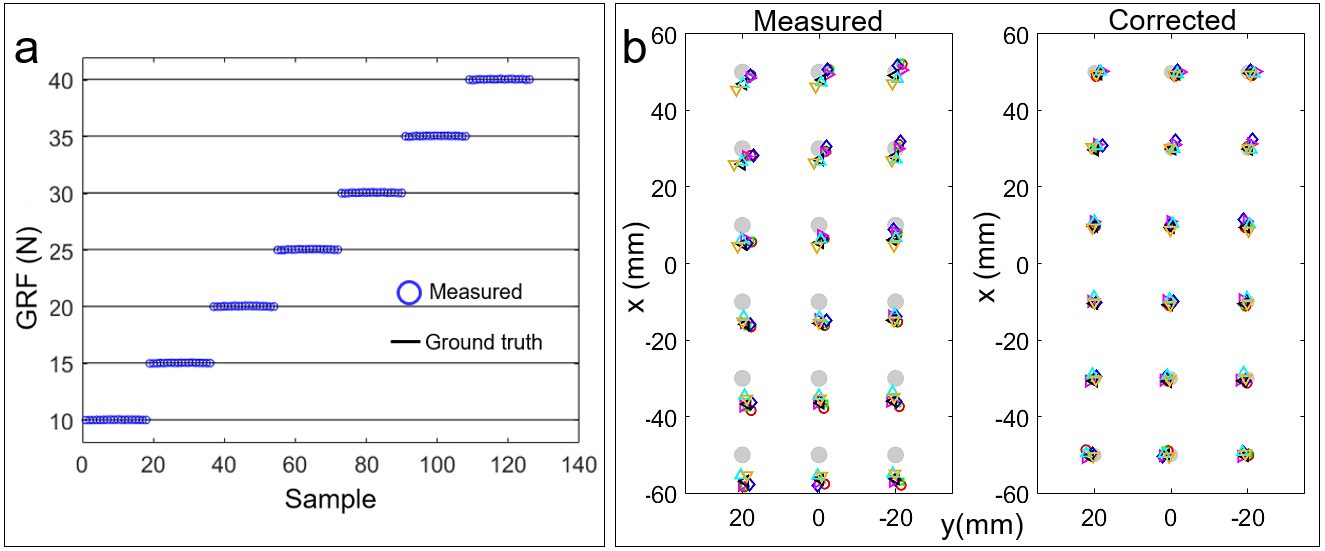}
\caption{Left shoe measurement results using manual calibration. (a) Measured GRF (blue circles) and their ground truths (black lines). (b) Measured CoP (left) and the corrected CoP (right). Colored markers represent measurements using different weights and grey circles are their corresponding ground truths.}
\label{ManualCalibrationResult}
\end{figure}

As introduced in \cite{han2021look}, our force-sensing shoes possess high accuracy in GRF measurement (Fig.~\ref{ManualCalibrationResult}a) but low accuracy in CoP measurements (Fig.~\ref{ManualCalibrationResult}b, left). To improve the CoP measurement accuracy, we apply seven different weights (1, 1.5, 2, 2.5, 3, 3.5, 4kg masses $\approx 18\%\text{, }27\%\text{, }37\%\text{, }46\%\text{, }55\%\text{, }64\%\text{, }73\%$ of the robot's mass) to the holes on the shoe's top plate using a 3-D printed weight support (Fig. \ref{shoecal}b shows three examples). Then, we utilize nonlinear least squares to minimize the error between measured CoPs and their corresponding ground truths, which are the known 2-D positions of the holes. 

Inspired by the force plate calibration method introduced in \cite{verkerke2005determining}, our manual calibration defines a corrected CoP, $[p_{x}, p_{y}]^{T}$, by summing the measured CoP $[p_{0x}, p_{0y}]^{T}$ and a correction term $[\Delta x, \Delta y]^{T}$:
\begin{align}
    \begin{bmatrix}
        p_{x} \\ p_{y}
    \end{bmatrix} = 
    \begin{bmatrix}
        p_{0x} \\ p_{0y}
    \end{bmatrix} +
    \begin{bmatrix}
        \Delta{x} \\ \Delta{y}
    \end{bmatrix}.
    \label{cor_term}
\end{align}
The correction terms are formulated as:
\begin{align}
    \begin{bmatrix}
       \Delta{x} \\\Delta{y}
    \end{bmatrix} =
    \begin{bmatrix}
        a_{1}p_{0x}^{2} + a_{2}p_{0x} + a_{3}p_{0y} + a_{4} + \sum_{i=1}^{4}m_{i}f_{i}\\
        b_{1}p_{0y}^{2} + b_{2}p_{0y} + b_{3}p_{0x} + b_{4} + \sum_{i=1}^{4}n_{i}f_{i}
    \end{bmatrix}, 
    \label{cor_detail}
\end{align}
which are designed first using a second-order polynomial of the measured CoP, $\mathbf{p_{0}}$, parameterized using $a_{i}$, $b_{i}$ ($i = 1\hdots4$). This part is used to improve the averaged measurement at each calibration hole. The correction terms also includes first-order terms of the force outputs $f_{i}$ parameterized using $m_{i}$ and $n_{i}$ $(i = 1\hdots4$, corresponding to each force output). This second part is for reducing the deviation at the same calibration spot for different applied weights.
Eventually, nonlinear least squares is used to minimize the error between the corrected CoP, ${\bf p}$ and the ground truth, ${\bf c}$:
\begin{align}
     \underset{{\bf \zeta}}{\text{argmin}} \quad J  = \sum_{k=1}^{N}||{\bf c}[k] - {\bf p}[k]||^{2} \label{manual_correction_term},
\end{align}
where $k$ is the sample index, $N$ is the sample number, and $\zeta = [a_{1},\hdots,a_{4},m_{1},\hdots, m_{4},$ $b_{1},\hdots,b_{4}, n_{1},\hdots,n_{4}]$ in (\ref{cor_detail}) are the optimization variables.
\vspace{-2pt}
\subsection{Shoe measurement accuracy}\label{error}
Mean absolute error (MAE) is used to quantify the CoP and GRF accuracy. The MAE of the GRF, $e_{G}$, is given by:
\begin{align}
&e_{G} = (\sum_{k=1}^{N}|f_{z}[k] - G_{\text{weight}}[k]|)/N \label{eGRF},
\end{align}
where $f_{z}$ is the measured GRF and $G_{\text{weight}}$ is the calibration weight. 
The MAE of CoP, $e_{C}$, is defined as:
\begin{align}
&e_{C} = (\sum_{k=1}^{N}||{\bf c}[k] - {\bf p}[k]||)/N \label{eCoP}, 
\end{align}
where ${\bf p}$ and ${\bf c}$ are the measured CoP and the ground truth.

The example of GRF and CoP measurements of the left shoe are shown in Fig.~\ref{ManualCalibrationResult}. The MAEs of both shoes are listed in Table~\ref{shoe_Manual_table}. According to the results, the measured GRF (Fig. \ref{ManualCalibrationResult}a, blue circles) almost lie perfectly on their corresponding ground truths (black lines) with only around 0.03N MAE (Table~\ref{shoe_Manual_table}); The corrected CoPs (Fig. \ref{ManualCalibrationResult}b, right, colored markers) become much closer to their corresponding ground truth (gray circles) compared with the initial measurements (left, colored markers). The accuracy of the corrected CoPs is improved about five times with only around 1 mm MAE (Table~\ref{shoe_Manual_table}). The results indicate that our manual calibration enables high CoP measurement accuracy. 
\begin{table}[H]
\caption{Manual Calibration (MAE)} 
\centering 
\begin{tabular}{|c|c|c|c|}
 \hline
Shoe & Measured GRF (N) & Measured CoP (mm) & Corrected CoP (mm)\\
 \hline
 Left  & 0.02 $\pm$ 0.01 & 4.53 $\pm$ 1.89 &  1.07 $\pm$ 0.62\\
\hline
 Right & 0.03 $\pm$ 0.03 & 3.51 $\pm$ 1.90 &  0.76 $\pm$ 0.35\\
\hline
\end{tabular}
\label{shoe_Manual_table}
\end{table}

\section{Shoe Self-Calibration}\label{AutonomousCalSection}
\subsection{Method overview}
When a humanoid robot moves slowly under a quasi-static assumption, its CoP equals the ground projection of its CoM \cite{kajita2014introduction}; Its normal GRF is simply the robot's weight. Ideally, the measured CoP and normal GRF using force sensors should equal their corresponding modeled values. Therefore, our self-calibration method collects modeled CoP, GRF and their corresponding force sensor outputs by moving the robot along planned trajectories in different double support configurations. Then the sensor parameters can be solved by minimizing the error between the sensor measurements and their modeled references using optimization.

\subsection{Data sampling strategy}
In order to initiate the self-calibration, the robot needs to first reach different foot configurations. Two options are considered to sample these configurations: \textbf{1.} \textit{The robot stands with single support and self-calibrate each shoe separately.} \textbf{2.} \textit{The robot stands with double support and self-calibrate both shoes together.} The former is ideal since it calibrates only four sensors of one shoe, reducing the potential of over-fitting by limiting the dimension of the optimization space. However, this option requires the robot to move its body with only one foot supporting its weight, resulting in excessive torque for smaller-sized humanoid robots with under-powered motors. Practically, our NAO robot platform encountered serious ankle overheating issues standing in single support during our initial testing. Therefore, the double support option is chosen for safety concerns. 

 To prevent overfitting for calibrating eight sensors in both feet, we collect sensor voltage data covering a sufficient range. Additionally, we ensure a large sample size. Two strategies are implemented to realize the above sampling results. \textbf{1}. We sample double supports with the position and orientation of one foot sufficiently far from the other (Section~\ref{double_support_sample}). \textbf{2}. We plan whole-body trajectory for each sampled double support such that the robot's CoP covers different areas of the sensing polygon (Section~\ref{CoP_Traj_sample}).
\begin{figure}[t!]
\centering
\includegraphics[width=0.9\linewidth]{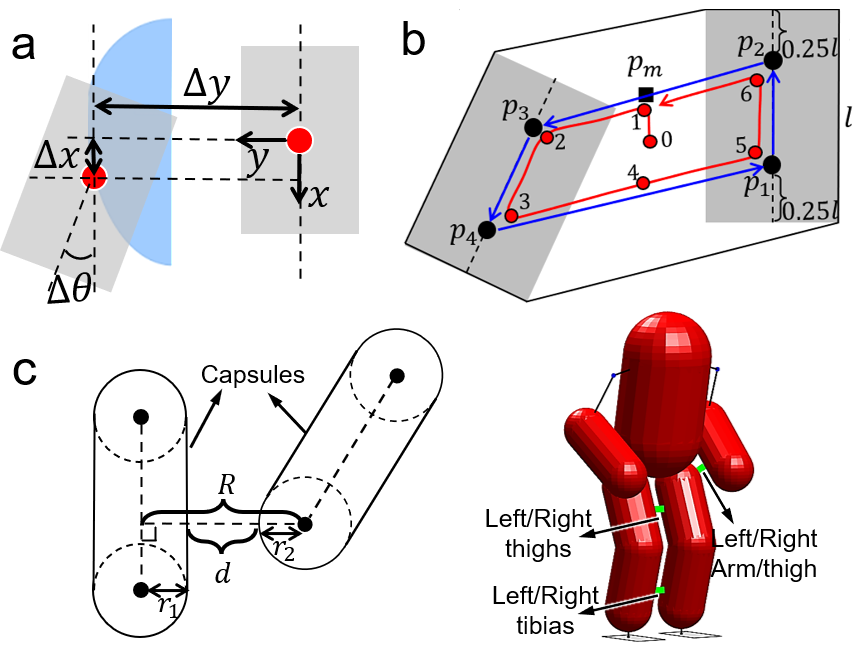}
\caption{(a) The configuration of the double support (grey rectangles) is defined by $\Delta x$, $\Delta y$ and $\Delta \theta$. Blue region is the factory allowable position of one foot relative to the other. (b) CoP trajectory generation. $p_{1}-p_{4}$ are the CoP landmarks. Blue and red curves are the desired and the planned CoP trajectories. point $0$ is the initial CoP position and $p_{m}$ is the initial CoP target. (c) left: the schematic showing the minimal distance between two 3-D capsules, where $r_{1}$,$r_{2}$ are the radii of the collision pairs shown in the right. $d$ and $R$ are the minimal distance between the capsules and between their centerlines.}
\label{traj_gen}
\end{figure}

\section{Motion Planning}
This section introduces the details for double support generation and whole-body motion planning.
\subsection{Double support configuration generation}\label{double_support_sample}
Each double support configuration is defined using the position and orientation of one shoe relative to another with three parameters: $[\Delta x, \Delta y$,$\Delta \theta]$ (Fig. \ref{traj_gen}a). We first discretize these parameters in the robot's factory allowable ranges (blue area) and then randomly sample in the discretized space. For each sampled double support configuration, a collision check is implemented to ensure no collision between the feet. The configuration distance between the feet of each sample must also be greater than a threshold, $d$, designed as:
\begin{align}
    d = w_{d}\sqrt{{\Delta{x}}^{2} + {\Delta{y}}^{2}} + w_{o}|\Delta \theta|,  
\end{align}
where $w_{d}$ and $w_{o}$ are the weights of the Cartesian distance and the relative orientation. Once the newly sampled double support passes both checks, it is stored for later execution.
\subsection{Whole-body trajectory generation}\label{CoP_Traj_sample}
{For each sampled double support, a whole-body trajectory is generated using trajectory optimization (Fig. \ref{traj_gen}b). First, four CoP landmarks are designed at the sensing polygon's centerlines and are 0.25-length away from the sensing polygon's front or rear edge ($p_{1} - p_{4}$, $l$ is the foot length). Then a trajectory planner generates a series of joint angle vectors, enabling the modeled CoP (red curve) to move counter-clockwise along the polygon contour constructed by the landmarks (blue curve). Initially, the CoP is driven towards one of the midpoints of the polygon contour ($p_{m}$) that is closest to the robot's initial CoP (point 0). The robot's arms are fixed in the trajectory, as they have little impact on CoP.}

We adopt the direct collocation technique \cite{tsang1975optimal} for trajectory optimization because it effectively solves trajectory including complex path constraints like our case. The optimization is formulated as:
\begin{align}
\underset{{\bf u},{\bf q}}{\text{minimize}} \quad & \sum_{i=1}^{N}(||{\bf c}[i] - {\bf c}_{\text{ref}}||^{2}_{Q_{c}} + ||{\bf u}[i] - {\bf u}[i-1]||^{2}_{Q_{u}}) \label{selfcal_cost}\\
\textrm{subject to:} \quad
&{\bf q}[i] = {\bf q}[i-1] + {\bf u}[i-1]  \quad (\text{state}) \label{selfcal_state_trans}\\
&{\bf c}[i] = \textbf{CoP}({\bf q}[i]) \in {\bf SP} \quad (\text{stability}) \label{selfcal_CoP}\\
&{\bf q}_{\text{min}}<{\bf q}[i]<{\bf q}_{\text{max}} \quad (\text{joint limit}) \label{selfcal_jointlim}\\
&{\bf d}[i] = \textbf{Dist}({\bf q}[i])> {\bf d}_{\text{min}} \quad (\text{collision}) \label{selfcal_collision}\\
&{\bf T}[i] = \text{TF}({\bf q}[i]) = {\bf T}_{0} \quad (\text{foot TF}) \label{selfcal_transformation}
\end{align}
\begin{algorithm}[t!] 
\caption{Trajectory planning} \label{traj_alg}
\begin{algorithmic}
\STATE Command robot to reach a sampled double support
\STATE Obtain robot's current body configuration
\STATE $n \gets 1$ \quad (initialize CoP landmark index: $n$)
\STATE $s \gets 0$ \quad (initialize planning step: $s$)
\STATE ${\bf d}[s] \gets ||{\bf c}[s] - {\bf l}[n]||$  (initialize CoP to landmark distance)
\WHILE{$n < N$}
    \STATE Run optimization \quad (\ref{selfcal_cost}--\ref{selfcal_transformation})
    \STATE $s \gets s+1$
    \STATE$ {\bf d}[s] \gets ||{\bf c}[s] - {\bf l}[n]||$ 
    \IF{${\bf d}[s] < r \quad \text{or} \quad {\bf d}[s-1] - {\bf d}[s] < 0$}
        \STATE $n \gets n+1$
    \ENDIF
\ENDWHILE
\end{algorithmic}
\end{algorithm}
where the system states are a series of joint angle vectors: ${\bf q} = [{\bf q_{1}},\hdots,{\bf q_{N}}]$, and the transition vectors: ${\bf u}=[{\bf u_{1}},\hdots,{\bf u_{N-1}}]$. The transition vectors connect consecutive states along the trajectory (\ref{selfcal_state_trans}). The cost function (\ref{selfcal_cost}) penalizes the distance between the modeled CoP ${{\bf c}[i]}$ and the CoP landmark ${\bf c_{\text{ref}}}$, driving the CoP trajectory towards the landmark. Moreover, the cost (\ref{selfcal_cost}) minimizes the state transition difference to smooth the trajectory. Several path constraints are imposed: the stability constraint (\ref{selfcal_CoP}) confines the modeled CoP inside the sensing polygon of the double support. The joint limit constraint (\ref{selfcal_jointlim}) determines the lower and upper bound of the states. The collision constraints (\ref{selfcal_collision}) prevent leg collision. 3-D capsules are used to approximate the robot's link geometry (Fig. \ref{traj_gen}c). The collision constraints set the minimal distance $d$ of the collision pairs (\ref{traj_gen}c, right) larger than a smaller threshold $d_{\text{min}}$. This minimal distance is calculated using the capsule centerline distance ($R$) to subtract the sum of their radii ($r_{1} + r_{2}$) (Fig. \ref{traj_gen}c, left). The foot transformation constraint (\ref{selfcal_transformation}) preserves the relative position and orientation between the feet for all the states in the trajectory. Finally, the optimization is solved using a nonlinear programming solver.

In practice, it is challenging for the solver to generate a complete trajectory directly connecting adjacent CoP landmarks in the sampled double supports (Fig. \ref{traj_gen}b). On the one hand, the convergence of the optimizer is sensitive to the number of states in the trajectory. On the other hand, a feasible trajectory may not exist once the robot's configuration gets closer to its singularity, which occurs when the CoP is close to its landmark. To ensure our algorithm runs autonomously, the planner is implemented in a model predictive control scheme: the optimizer continuously generates a small portion of the trajectory ahead of time, moving the CoP towards its landmark. Once the CoP is close enough to the landmark, or the CoP trajectory cannot proceed more towards the landmark, a new landmark is updated. The detailed implementation of the planner is presented in Algorithm~\ref{traj_alg} and the snapshots of the generated whole-body trajectories are shown in Fig. \ref{autonomous_result1} and \ref{autonomous_result2}.
\section{Implementation Details}
\subsection{Data preparation}
To implement the self-calibration algorithm, we sampled five different double supports and then generated their corresponding whole-body trajectories. Among these datasets, three are randomly chosen for training (Fig. \ref{autonomous_result1}). The other two (Fig. \ref{autonomous_result2}) are used for evaluation.
\begin{figure}[t!]
\centering
\includegraphics[width=1.0\linewidth]{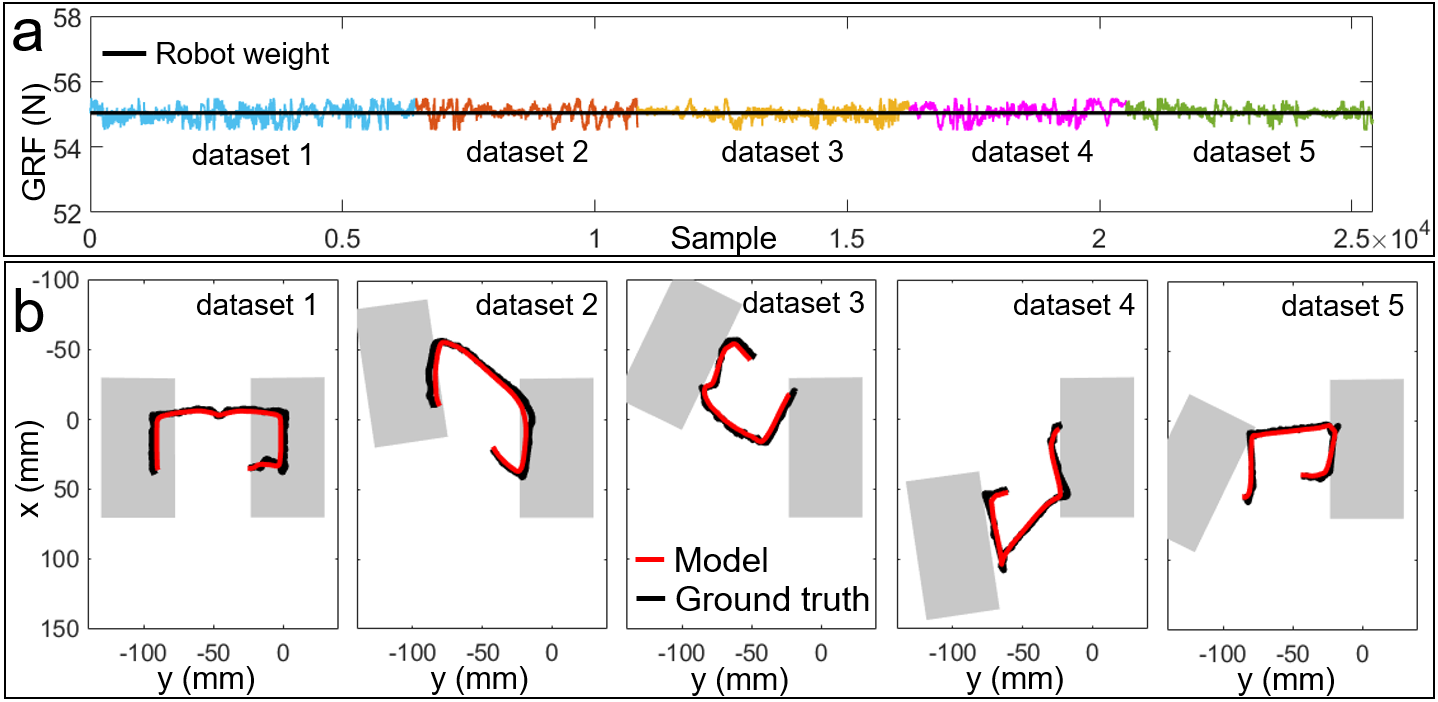}
\caption{(a) Quasi-static assumption evaluation. Black line represents the robot's weight and the colored lines represent the measured GRFs using manual calibration for the five sampled datasets. (b) Comparison between modeled CoPs and their corresponding ground truths obtained by manual calibration. Grey rectangles show the sensing polygon of each foot. Black and red curves show the CoP ground truth and the modeled CoP.}
\label{quasi_check}
\end{figure}
\begin{table}
\caption{GRF $\&$ CoP Mean Absolute Error} 
\centering 
\begin{tabularx}{0.48\textwidth} { 
   |>{\centering\arraybackslash}X | 
    >{\centering\arraybackslash}X |
    >{\centering\arraybackslash}X |
    >{\centering\arraybackslash}X |
    >{\centering\arraybackslash}X |
    >{\centering\arraybackslash}X |}
 \hline
 & data 1 & data 2 &  data 3 & data 4 & data 5\\
 \hline
GRF \quad (N) & 0.27 $\pm$ 0.22 & 0.27 $\pm$ 0.20 & 0.12 $\pm$ 0.16 & 0.25 $\pm$ 0.21 & 0.31 $\pm$ 0.16 \\
\hline
CoP \quad (mm) & 2.98 $\pm$ 1.62 & 2.72 $\pm$ 1.37 & 2.20 $\pm$ 1.19 & 2.50 $\pm$ 1.71 & 2.01 $\pm$ 1.56  \\
\hline
\end{tabularx}
\label{quasi_check_table}
\end{table}
\subsection{Modeling reliability}\label{CoPModelAccuracy}
For the self-calibration to work effectively, two assumptions need to be validated. \textbf{Assumption 1:} The robot's slow movement can be modeled as a quasi-static system. \textbf{Assumption 2:} The modeled CoP is sufficiently close to the ground truth. To test these two assumptions, we manually calibrated the shoes and use their high-accurate measurements (see Table~\ref{shoe_Manual_table}) as ground truths to evaluate the modeled results. For Assumption 1, we compared the measured normal GRF using manual calibration and the robot's weight. For Assumption 2, we compared the modeled CoP and the measured CoP using manual calibration. The results show that the measured GRF (Fig. \ref{quasi_check}a, colored curves) slightly oscillates around the ground truth (black line) with only around 0.3 N MAE (Table~\ref{quasi_check_table}), indicating that the robot's slow movement is close enough to quasi-static condition. In addition, the modeled CoP trajectories (Fig. \ref{quasi_check}b, red curves) are close to the measured references (black curves) with less than 3 mm MAE (Table~\ref{quasi_check_table}), indicating that the modeling is with decent fidelity. The experimental results prove that both assumptions are practically applicable.

\subsection{Center of pressure modeling}
Under quasi-static assumption, the modeled CoP equals the projection of the modeled CoM of the robot. With a priori known mass properties of the robot's body components, the CoM can be modeled by solving whole body forward kinematics using joint angle data obtained by encoders.

\subsection{Initial sensor parameters estimation} \label{InitGuess}
The self-calibration utilizes nonlinear least squares to determine the sensor parameters, requiring a reasonable initial guess. Since all load cells are of the same type, we apply the same initial guess $[c_{0}, d_{0}]$ (see \ref{load-cell-cal} for load cell modeling) to all the sensors. Assuming the measured GRF and CoP using initial guess approximately equal their modeled references, the following equations hold:
\begin{align}
    \begin{bmatrix}
    G_{r}\\
    c_{x} \\
    c_{y} 
    \end{bmatrix} \approx 
    \begin{bmatrix}
    \sum_{i=1}^{8}(c_{0}S[i]+d_{0})\\
    \sum_{i=1}^{8}(c_{0}S[i]+d_{0})t_{x}[i]/\sum_{i=1}^{8}(c_{0}S[i]+d_{0})\\
   \sum_{i=1}^{8}(c_{0}S[i]+d_{0})t_{y}[i]/\sum_{i=1}^{8}(c_{0}S[i]+d_{0})
    \end{bmatrix}. 
\end{align}
The left side consists of the robot's weight, $G_{r}$, and the modeled CoP, $[c_{x}, c_{y}]^{T}$. The right side consists of the measured GRF and CoP, where $i$ is the sensor index, $S$ and $(t_{x},t_{y})$ are the sensor voltage output and location. We can rearrange the equations to include more training samples:
\begin{align} \label{initial_guess_equ}
    \begin{bmatrix}
    \vdots\\
    c_{x}[k]G_{r} \\
    c_{y}[k]G_{r} \\
    \vdots
    \end{bmatrix} \approx 
    \begin{bmatrix}
    \vdots & \vdots\\
    \sum_{i=1}^{8}S[k][i]t_{x}[i]&\sum_{i=1}^{8}t_{x}[i]\\
    \sum_{i=1}^{8}S[k][i]t_{y}[i]&\sum_{i=1}^{8}t_{y}[i]\\
    \vdots & \vdots
    \end{bmatrix} 
    \begin{bmatrix}
    c_{0}\\
    d_{0}
    \end{bmatrix}
\end{align}
where $k$ is the sample number. Therefore, the optimal initial guess $(c_{0},d_{0})$ can be solved by least squares regression.
\begin{figure*}[t!]
\centering
\includegraphics[width=0.85\textwidth]{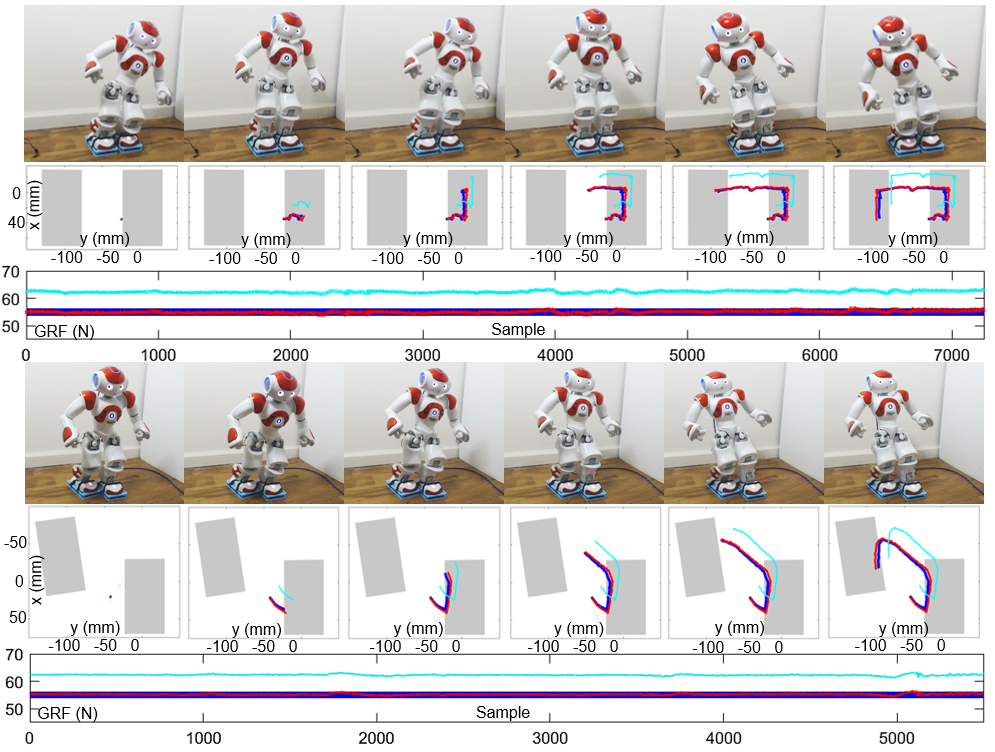}
\caption{Two example training datasets. Top, middle and bottom sections for each dataset show the evolution of the robot's motion, its CoP and GRF. Teal, blue and red lines show the measurements using the initial guess, the ground truth and the self-calibration.}
\label{autonomous_result1}
\end{figure*}
\subsection{Sensor parameter identification}\label{loadcellrecovery}
\subsubsection{Loadcell parameter identification}
Given training samples consisting of sensor voltage outputs and their corresponding modeled GRFs and CoPs (Fig.~\ref{autonomous_result1}), the load cells' parameters are determined by minimizing the differences between these corresponding pairs using nonlinear least squares (NLS). The optimization variables ${\bf \zeta} = [(c_{1},d_{1}),\hdots,(c_{8},d_{8})]$ are the parameters of the load cells' scaling factors and offsets in (\ref{affine}). The NLS is formulated as:
 \begin{align}\label{GRF_recovery} 
     \underset{{\bf \zeta}}{\text{argmin}}\quad J  = \sum_{k=1}^{N}(|n[k] - G_{\text{robot}}|^{2}_{{\bf w_{n}}}+ \\ \nonumber
     ||{\bf c}[k] - {\bf c}_{m}[k]||^{2}_{{\bf w_{c}}}) + ||{\bf \zeta}-{\bf \zeta_{0}}||^{2}_{{\bf w_{\zeta}}}, 
\end{align}
where $k$ is the number of training samples, ${n}$ and ${\bf c}$ are measured GRF and CoP; ${\bf c_{m}}$ is the modeled CoP; $G_{\text{robot}}$ is the robot's weight; ${\bf \zeta_{0}}$ is the initial guess solved by (\ref{initial_guess_equ}). $w_{n}$, $w_{c}$ and $w_{\zeta}$ are the weights for the cost terms. The regulation term in the cost is designed to avoid overfitting. 
\subsubsection{CoP measurement calibration}
Similar to the manual calibration method (see \ref{shoe_cop_cal_section}), we further improve the CoP measurement by first defining a corrected CoP:
\begin{align}
  \quad {\bf c}_{\text{aug}} = \frac{({\bf c}_{L}+\Delta {\bf s}_{L})n_{L}+({\bf c}_{R}+\Delta {\bf s}_{R})n_{R}}{n_{L}+n_{R}},
\end{align}
where $c_{L}$, $c_{R}$, $n_{L}$ and $n_{R}$ are the measured CoP and GRF using sensor parameters for the left and right feet solved by (\ref{GRF_recovery}). $\Delta s_{L}$ and $\Delta s_{R}$ are the correction terms applied to the left and right feet defined in (\ref{cor_detail}). Then we use NLS to minimize the error between corrected and modeled CoPs over the parameters of the correction terms:
\begin{align}
     \underset{{\bf \zeta}}{\text{argmin}} \quad J  = \sum_{k=1}^{N}(||{\bf c}_{\text{aug}}[k] - {\bf c_{m}}[k]||^{2}),
\end{align}
where $k$ is the sample number, ${\bf c}_{m}$ is the modeled CoP and $\bf \zeta$ are the parameters of the correction terms defined in (\ref{cor_detail}). 
\begin{figure*}[t!]
\centering
\includegraphics[width=0.84\textwidth]{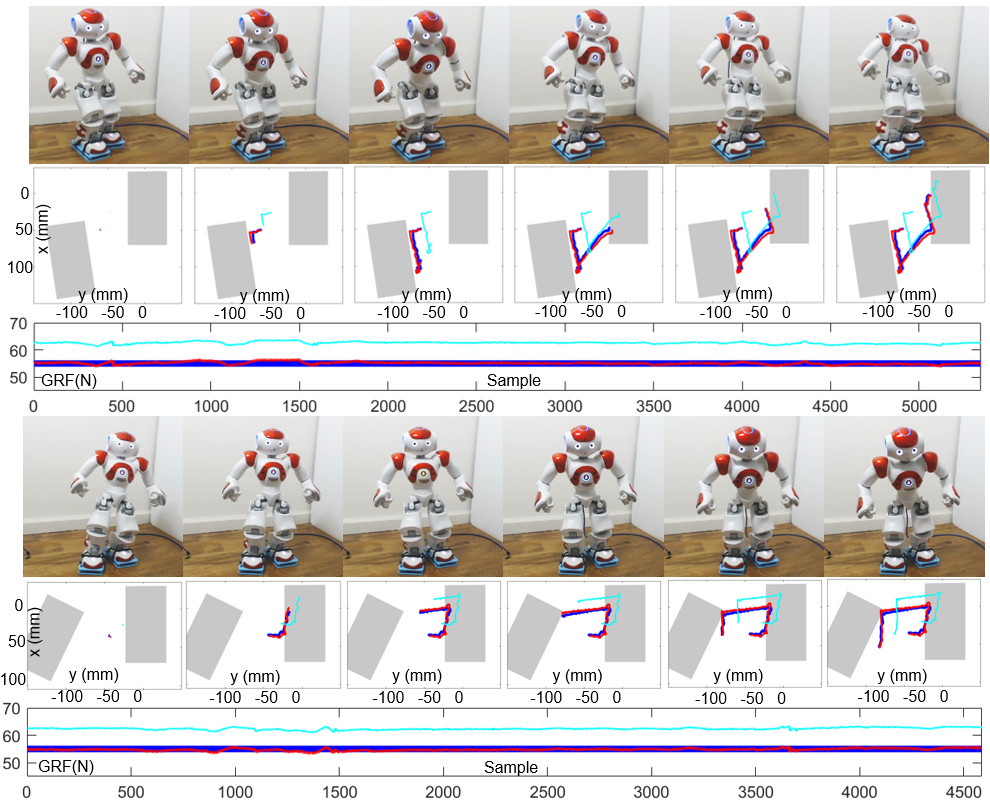}
\caption{Two testing datasets. Top middle and bottom sections for each dataset show the evolution of the robot's motion, its CoP and GRF. Teal, blue and red lines show the measurements using the initial guess, the ground truth and the self-calibration.}
\label{autonomous_result2}
\end{figure*}
\begin{figure}[t!]
\centering
\includegraphics[width=1.0\linewidth]{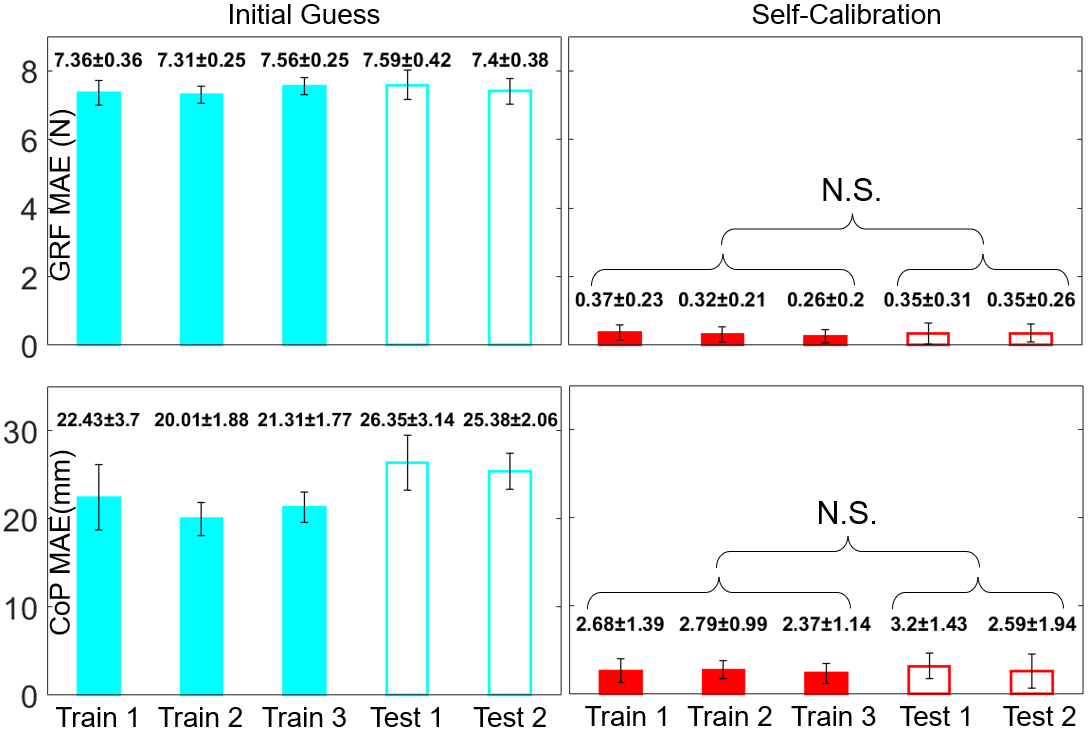}
\caption{Top and bottom show the MAE of the measured GRF and CoP using initial guess (left) and self-calibration (right). Filled and empty bars show the training and testing results. N.S means no statistical difference.}
\label{self_result_fig}
\end{figure}
\section{Results}
The shoe measurements using self-calibrated parameters and the corresponding ground truths acquired by manually calibrated parameters for the training and testing datasets are shown in Fig. \ref{autonomous_result1} and \ref{autonomous_result2}. The MAE of the measurements using initial guess (Section~\ref{InitGuess}) and the measurements using self-calibration are presented in Fig. \ref{self_result_fig}. The results show that GRF and CoP measurements using initial guess (Fig. \ref{autonomous_result1} and \ref{autonomous_result2}, teal lines) are far away from their ground truths (blue lines), with about 7.5 N MAE for GRF and 23 mm MAE for CoP (Fig. \ref{self_result_fig}, left). By contrast, the measurement using self-calibration (Fig. \ref{autonomous_result1} and \ref{autonomous_result2}, red lines) are much closer to their corresponding ground truths (blue lines) with only approximate 0.35 N MAE for GRF and 2.8 mm MAE for CoP. The self-calibration improves the measurements using initial guess by approximately ten times for CoP (Fig. \ref{self_result_fig}, 
bottom) and 20 times for GRF (Fig. \ref{self_result_fig}, top). In addition, statistical tests show that the training (Fig. \ref{self_result_fig}, right, filled bars) and testing (empty bars) MAEs are not significantly different, indicating no overfitting takes place. The overall results demonstrate that our self-calibration approach can effectively determine sensor parameters without any manual intervention and without needing initial sensor information.

\section{Discussion}
This letter presents a novel self-calibration method for humanoid robots to determine the sensor parameters of their foot force-sensing modules autonomously. Our method does not require any manual intervention. Therefore, this approach enables a calibration process without the need of detaching the device from the robot or disassembling the device. In addition, this method can also be applied for sensor correction in long-term autonomous tasks. Although the self-calibration approach is demonstrated on our force-sensing shoes designed for a NAO humanoid robot, it can theoretically be applied to other foot-sensing modules for both smaller and larger humanoids, using similar planning and optimization procedures as ours.

It is observable that our planned CoP trajectories in the training and testing datasets do not cover the edge of the sensing polygon (Fig. \ref{autonomous_result1} and \ref{autonomous_result2}). This is due to the limitation of our smaller-sized robot platform. During our initial testing, we discovered that when the robot's CoP is near the edge of the sensing polygon, its major supporting leg almost always suffered from excessive torque. Therefore, we designed CoP trajectory to not cover the edge of the sensing polygon for safety concerns.

Compared with the ground truth produced by manually calibrated sensor parameters (Table~\ref{shoe_Manual_table}), the measurement using self-calibrated parameters is slight off (Fig.~\ref{self_result_fig}, right), which is likely caused by the following issues: 1. the robot's movement can never reach fully quasi-static since the acceleration always exists during the movement of the robot (Fig. \ref{quasi_check}a and Table~\ref{quasi_check_table}). 2. The modeled CoP does not perfectly equal the ground truth (Fig. \ref{quasi_check}b and Table~\ref{quasi_check_table}). These problems lead to fitting errors in the optimization process.

\section{Acknowledgement}
This work was supported by NUS Startup grants R-265-000-665-133, R-265-000-665-731, Faculty Board account C-265-000-071-001, and the National Research Foundation, Singapore under its Medium Sized Center for Advanced Robotics Technology Innovation R-261-521-002-592.


\bibliographystyle{IEEEtran}
\bibliography{ICRA.bib}

\begin{thebibliography}{10}
\providecommand{\url}[1]{#1}
\csname url@samestyle\endcsname
\providecommand{\newblock}{\relax}
\providecommand{\bibinfo}[2]{#2}
\providecommand{\BIBentrySTDinterwordspacing}{\spaceskip=0pt\relax}
\providecommand{\BIBentryALTinterwordstretchfactor}{4}
\providecommand{\BIBentryALTinterwordspacing}{\spaceskip=\fontdimen2\font plus
\BIBentryALTinterwordstretchfactor\fontdimen3\font minus
  \fontdimen4\font\relax}
\providecommand{\BIBforeignlanguage}[2]{{%
\expandafter\ifx\csname l@#1\endcsname\relax
\typeout{** WARNING: IEEEtran.bst: No hyphenation pattern has been}%
\typeout{** loaded for the language `#1'. Using the pattern for}%
\typeout{** the default language instead.}%
\else
\language=\csname l@#1\endcsname
\fi
#2}}
\providecommand{\BIBdecl}{\relax}
\BIBdecl

\bibitem{vukobratovic2004zero}
M.~Vukobratovi{\'c} and B.~Borovac, ``Zero-moment point—thirty five years of
  its life,'' \emph{International journal of humanoid robotics}, vol.~1,
  no.~01, pp. 157--173, 2004.

\bibitem{tsuichihara2011sliding}
S.~Tsuichihara, M.~Koeda, S.~Sugiyama, and T.~Yoshikawa, ``A sliding walk
  method for humanoid robots using zmp feedback control,'' in \emph{2011 IEEE
  International Conference on Robotics and Biomimetics}.\hskip 1em plus 0.5em
  minus 0.4em\relax IEEE, 2011, pp. 275--280.

\bibitem{nakaura2002balance}
S.~Nakaura, M.~Sampei \emph{et~al.}, ``Balance control analysis of humanoid
  robot based on zmp feedback control,'' in \emph{IEEE/RSJ International
  Conference on Intelligent Robots and Systems}, vol.~3.\hskip 1em plus 0.5em
  minus 0.4em\relax IEEE, 2002, pp. 2437--2442.

\bibitem{ghassemi2014push}
P.~Ghassemi, M.~T. Masouleh, and A.~Kalhor, ``Push recovery for nao humanoid
  robot,'' in \emph{2014 Second RSI/ISM International Conference on Robotics
  and Mechatronics (ICRoM)}.\hskip 1em plus 0.5em minus 0.4em\relax IEEE, 2014,
  pp. 035--040.

\bibitem{hawley2016external}
L.~Hawley and W.~Suleiman, ``External force observer for medium-sized humanoid
  robots,'' in \emph{2016 IEEE-RAS 16th International Conference on Humanoid
  Robots (Humanoids)}.\hskip 1em plus 0.5em minus 0.4em\relax IEEE, 2016, pp.
  366--371.

\bibitem{piperakis2018nonlinear}
S.~Piperakis, M.~Koskinopoulou, and P.~Trahanias, ``Nonlinear state estimation
  for humanoid robot walking,'' \emph{IEEE Robotics and Automation Letters},
  vol.~3, no.~4, pp. 3347--3354, 2018.

\bibitem{han2020can}
Y.~Han, R.~Li, and G.~S. Chirikjian, ``Can i lift it? humanoid robot reasoning
  about the feasibility of lifting a heavy box with unknown physical
  properties,'' in \emph{2020 IEEE/RSJ International Conference on Intelligent
  Robots and Systems (IROS)}.\hskip 1em plus 0.5em minus 0.4em\relax IEEE,
  2020, pp. 3877--3883.

\bibitem{shigematsu2018lifting}
R.~Shigematsu, S.~Komatsu, Y.~Kakiuchi, K.~Okada, and M.~Inaba, ``Lifting and
  carrying an object of unknown mass properties and friction on the head by a
  humanoid robot,'' in \emph{2018 IEEE-RAS 18th International Conference on
  Humanoid Robots (Humanoids)}.\hskip 1em plus 0.5em minus 0.4em\relax IEEE,
  2018, pp. 1--9.

\bibitem{takenaka2006control}
T.~Takenaka, ``The control system for the honda humanoid robot,'' \emph{Age and
  ageing}, vol.~35, no. suppl\_2, pp. ii24--ii26, 2006.

\bibitem{koch2014optimization}
K.~H. Koch, K.~Mombaur, O.~Stasse, and P.~Soueres, ``Optimization based
  exploitation of the ankle elasticity of hrp-2 for overstepping large
  obstacles,'' in \emph{2014 IEEE-RAS International Conference on Humanoid
  Robots}.\hskip 1em plus 0.5em minus 0.4em\relax IEEE, 2014, pp. 733--740.

\bibitem{kajita2014introduction}
S.~Kajita, H.~Hirukawa, K.~Harada, and K.~Yokoi, \emph{Introduction to humanoid
  robotics}.\hskip 1em plus 0.5em minus 0.4em\relax Springer, 2014, vol. 101.

\bibitem{fujimoto1998attitude}
Y.~Fujimoto and A.~Kawamura, ``Attitude control experiments of biped walking
  robot based on environmental force interaction,'' in \emph{AMC'98-Coimbra.
  1998 5th International Workshop on Advanced Motion Control. Proceedings (Cat.
  No. 98TH8354)}.\hskip 1em plus 0.5em minus 0.4em\relax IEEE, 1998, pp.
  70--75.

\bibitem{gouaillier2009mechatronic}
D.~Gouaillier, V.~Hugel, P.~Blazevic, C.~Kilner, J.~Monceaux, P.~Lafourcade,
  B.~Marnier, J.~Serre, and B.~Maisonnier, ``Mechatronic design of nao
  humanoid,'' in \emph{2009 IEEE International Conference on Robotics and
  Automation}.\hskip 1em plus 0.5em minus 0.4em\relax IEEE, 2009, pp. 769--774.

\bibitem{takahashi2005high}
Y.~Takahashi, K.~Nishiwaki, S.~Kagami, H.~Mizoguchi, and H.~Inoue, ``High-speed
  pressure sensor grid for humanoid robot foot,'' in \emph{2005 IEEE/RSJ
  International Conference on Intelligent Robots and Systems}.\hskip 1em plus
  0.5em minus 0.4em\relax IEEE, 2005, pp. 3909--3914.

\bibitem{son2015development}
B.~J. Son, Y.~S. Baek, and J.~H. Kim, ``Development of foot modules of an
  exoskeleton equipped with multiple sensors for detecting walking phase and
  intent,'' in \emph{Applied Mechanics and Materials}, vol. 752.\hskip 1em plus
  0.5em minus 0.4em\relax Trans Tech Publ, 2015, pp. 1016--1021.

\bibitem{hollinger2006evaluation}
A.~Hollinger and M.~M. Wanderley, ``Evaluation of commercial force-sensing
  resistors,'' in \emph{Proceedings of the International Conference on New
  Interfaces for Musical Expression, Paris, France}.\hskip 1em plus 0.5em minus
  0.4em\relax Citeseer, 2006, pp. 4--8.

\bibitem{shayan2019design}
A.~M. Shayan, A.~Khazaei, A.~Hamed, and M.~T. Masouleh, ``Design and
  development of a pressure-sensitive shoe platform for nao h25,'' in
  \emph{2019 7th International Conference on Robotics and Mechatronics
  (ICRoM)}.\hskip 1em plus 0.5em minus 0.4em\relax IEEE, 2019, pp. 223--228.

\bibitem{kwon2011fabrication}
H.-J. Kwon, J.-H. Kim, D.-K. Kim, and Y.-H. Kwon, ``Fabrication of four-point
  biped robot foot module based on contact-resistance force sensor and its
  evaluation,'' \emph{Journal of mechanical science and technology}, vol.~25,
  no.~2, p. 543, 2011.

\bibitem{han2021look}
Y.~Han, R.~Li, and G.~S. Chirikjian, ``Look at my new blue force-sensing
  shoes!'' in \emph{2021 IEEE International Conference on Robotics and
  Automation (ICRA)}.\hskip 1em plus 0.5em minus 0.4em\relax IEEE, 2021, pp.
  2891--2896.

\bibitem{almeida2018novel}
L.~Almeida, V.~Santos, and F.~Silva, ``A novel wireless instrumented shoe for
  ground reaction forces analysis in humanoids,'' in \emph{2018 IEEE
  International Conference on Autonomous Robot Systems and Competitions
  (ICARSC)}.\hskip 1em plus 0.5em minus 0.4em\relax IEEE, 2018, pp. 36--41.

\bibitem{suwanratchatamanee2009haptic}
K.~Suwanratchatamanee, M.~Matsumoto, and S.~Hashimoto, ``Haptic sensing foot
  system for humanoid robot and ground recognition with one-leg balance,''
  \emph{IEEE Transactions on Industrial Electronics}, vol.~58, no.~8, pp.
  3174--3186, 2009.

\bibitem{NAOFSR}
``Nao robot's force sensing resistor introduction page,''
  \url{http://doc.aldebaran.com/2-5/family/robots/fsr_robot.html?highlight=fsr}.

\bibitem{verkerke2005determining}
G.~J. Verkerke, A.~Hof, W.~Zijlstra, W.~Ament, and G.~Rakhorst, ``Determining
  the centre of pressure during walking and running using an instrumented
  treadmill,'' \emph{Journal of biomechanics}, vol.~38, no.~9, pp. 1881--1885,
  2005.

\bibitem{tsang1975optimal}
T.~Tsang, D.~Himmelblau, and T.~F. Edgar, ``Optimal control via collocation and
  non-linear programming,'' \emph{International Journal of Control}, vol.~21,
  no.~5, pp. 763--768, 1975.

\end{thebibliography}

\end{document}